\documentclass[preprint,12pt]{elsarticle}

\usepackage{amssymb}
\usepackage{amsmath}
\usepackage{multirow}
\usepackage{graphicx}
\usepackage{subcaption}
\usepackage{float}
\usepackage{hyperref}

\journal{arXiv}

\begin{document}

\begin{frontmatter}



\title{Non-linear Analysis Based ECG Classification of Cardiovascular Disorders}


\author[label1]{Suraj Kumar Behera}
\author[label1]{Debanjali Bhattacharya}
\author[label1]{Ninad Aithal}
\author[label1]{Neelam Sinha}
\ead{neelam@cbr-iisc.ac.in
}
\affiliation[label1]{organization={Center for Brain Research, Indian Institute of Science (IISc) Campus},
            addressline={CV Raman Avenue}, 
            city={Bangalore},
            postcode={560012}, 
            state={Karnataka},
            country={India}
            }

\begin{abstract}
Multi-channel ECG-based cardiac disorders detection has an impact on cardiac care and treatment. Limitations of existing methods included variation in ECG waveforms due to the location of electrodes, high non-linearity in the signal, and amplitude measurement in millivolts. The present study reports a non-linear analysis-based methodology that utilizes Recurrence plot visualization. The patterned occurrence of well-defined structures, such as the QRS complex, can be exploited effectively using Recurrence plots. This Recurrence-based method is applied to the publicly available Physikalisch-Technische Bundesanstalt (PTB) dataset from PhysioNet database, where we studied four classes of different cardiac disorders (Myocardial infarction, Bundle branch blocks, Cardiomyopathy, and Dysrhythmia) and healthy controls, achieving an impressive classification accuracy of 100\%. Additionally, t-SNE plot visualizations of the latent space embeddings derived from Recurrence plots and Recurrence Quantification Analysis features reveal a clear demarcation between the considered cardiac disorders and healthy individuals, demonstrating the potential of this approach.
\end{abstract}




\end{frontmatter}



\section{INTRODUCTION}
\label{sec1}

Cardiovascular disorders (CVDs) are the leading causes of mortality globally, significantly impacting public health and driving up healthcare costs \cite{[1],[2]}. There are various types of CVDs. For example, Bundle branch block results from a disruption in electrical signals within the heart, causing abnormalities such as widened QRS complexes and altered electrocardiographic vectors. While the disease itself may not directly correlate with cardiovascular risk factors, the presence of a right bundle branch block can indicate a higher risk of mortality in conditions like myocardial infarction, heart failure, and various heart block types \cite{[3]}. Cardiomyopathy, also known as heart failure, impairs the heart's pumping ability, leading to irregular heartbeats and potentially severe complications. Dysrhythmia, or arrhythmia, involves irregular heart rhythms that can damage the heart or other organs, increasing the risk of stroke, heart failure, or cardiac arrest. Myocardial infarction, commonly referred to as a heart attack, occurs due to a critical reduction or cessation of blood flow to the heart muscle, making it a prevalent cause of death despite its relatively low global incidence.
Early detection of cardiac disorders through ECG signals is vital for reducing mortality rates. Clinically, diagnosing cardiac conditions involves examining deviations in the 12-lead ECG signals, particularly in the P, QRS, and T waveforms. However, variations in these waveforms can arise from the electrode placement, making detection challenging. The amplitude of ECG signals, measured in millivolts, further complicates the process due to its nonlinear nature \cite{[4]}. However, manually analyzing extensive amounts of ECG data is time-consuming and labor-intensive. An automated approach to ECG data analysis can drastically reduce detection time, improving early diagnosis and treatment outcomes. This is crucial for preventing disease progression, minimizing complications, and enhancing long-term health.
Recent studies have widely utilized various neural network models, such as convolutional neural networks \cite{[7],[8],deb2} and few-shot learning techniques \cite{[9]}, for classifying 1D signals. Additionally, several studies have transformed 1D signals into 2D images using techniques like Gramian Angular Fields, Markov Transition Fields, and cellular automata to improve the understanding of data dynamics and enhance classification performance \cite{[10],[11],[12],[13],[15],deb1,deb3}.

In contrast to existing approaches, the present study explores the applicability of \textit{Recurrence plots} for classifying various CVDs using ECG data. Recurrence plots offer a sophisticated method for nonlinear data analysis by visualizing a square matrix that captures the recurrence of states within a dynamical system. This technique is versatile, finding applications in diverse fields such as financial markets and cardiology, where it can reveal hidden patterns and structural changes in time signals \cite{[16]}\cite{[17]}. Recurrence plots are valuable for uncovering signatures of predictability, stationarity, fractal behavior, chaos, and noise. Previous studies have examined cardiovascular signals as sequences of deterministic trajectories interrupted by stochastic pauses (terminal dynamics), and have used recurrence quantification analysis (RQA) to assess such complex, nondeterministic behaviors. For example, Yafei Kang \cite{[19]} employed RQA to evaluate the duration, predictability, and complexity of periodic processes in time series, while Zbilut J.P. \cite{[18]} used cross-recurrence quantification analysis of ECG signals, followed by discrete wavelet transform, to investigate cardiac disorders.
 
The novelty of this study lies in the utilization of autoencoder latent-space embedding of Recurrence plots and RQA to classify different CVDs. This approach combines the advantages of autoencoders with Recurrence plots, leveraging latent-space representations to enhance the analysis of non-linear dynamics in ECG signals. 
By embedding Recurrence plots into the latent space of an autoencoder, the study aims to improve the extraction and quantification of complex patterns and features, thereby providing a more robust and accurate classification of various cardiac conditions. This method offers a promising direction for advanced ECG analysis and the early detection of CVDs through sophisticated non-linear data analysis techniques.

\section{DATASET DESCRIPTION}
\label{sec2}

\begin{figure}[H]
    \centering
    \begin{subfigure}[b]{0.45\linewidth}
        \centering
        \includegraphics[width=\linewidth]{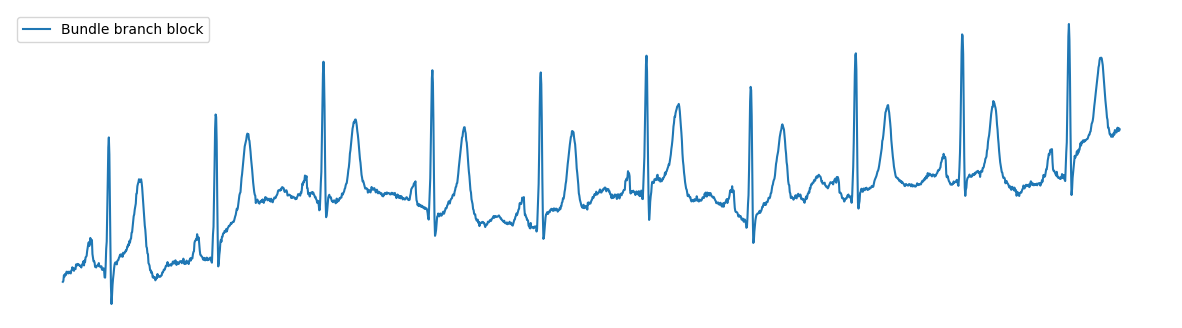}
        \caption{ECG Plot for Bundle branch block}
        \label{fig:a}
    \end{subfigure}
    \hfill
    \begin{subfigure}[b]{0.45\linewidth}
        \centering
        \includegraphics[width=\linewidth]{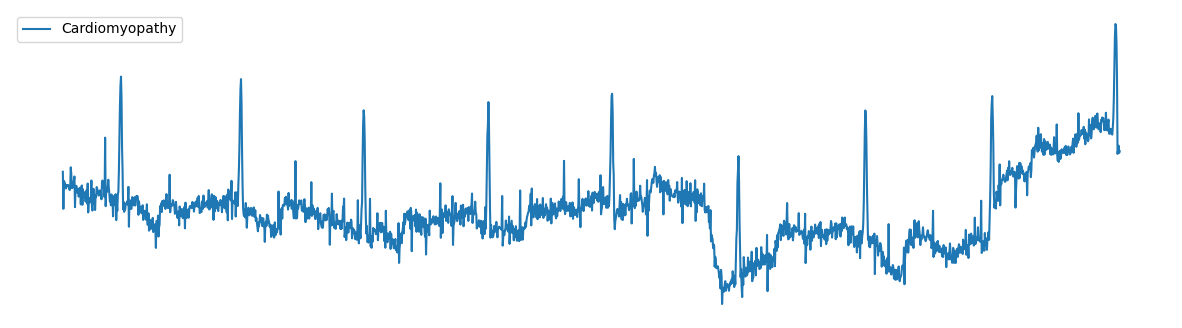}
        \caption{ECG Plot for Cardiomyopathy}
        \label{fig:b}
    \end{subfigure}
    \hfill
    \begin{subfigure}[b]{0.45\linewidth}
        \centering
        \includegraphics[width=\linewidth]{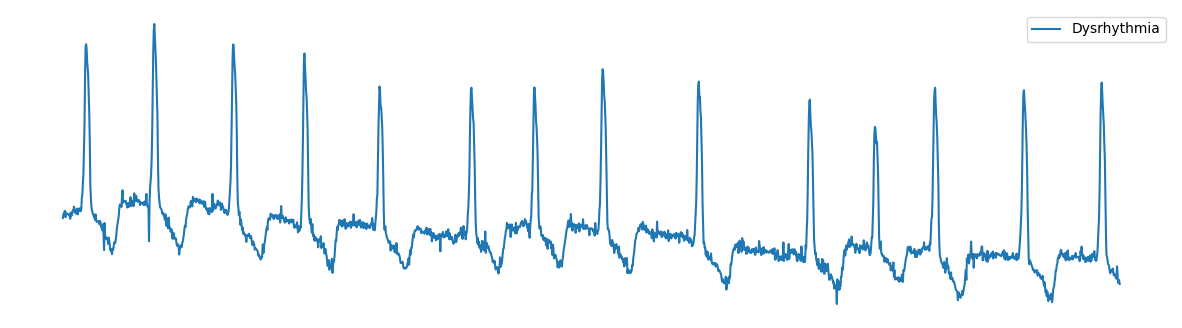}
        \caption{ECG Plot for Dysrhythmia }
        \label{fig:c}
    \end{subfigure}
    \hfill
    \begin{subfigure}[b]{0.45\linewidth}
        \centering
        \includegraphics[width=\linewidth]{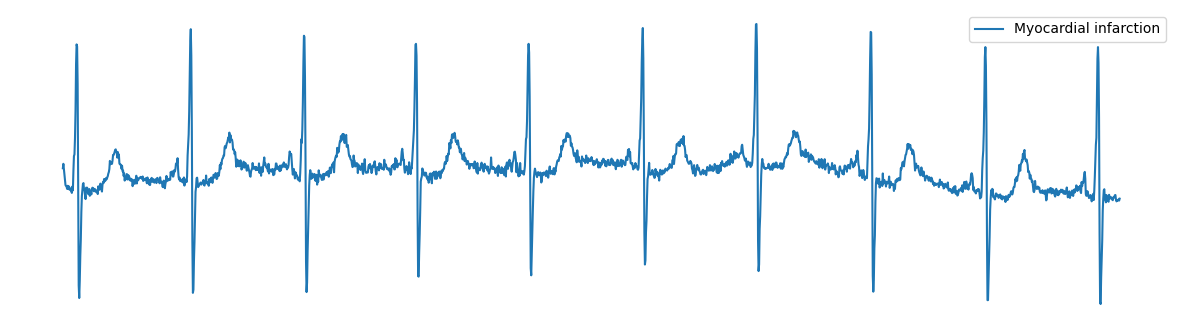}
        \caption{ ECG Plot for Myocardial infarction}
        \label{fig:d}
    \end{subfigure}
    \hfill
    \begin{subfigure}[b]{0.45\linewidth}
        \centering
        \includegraphics[width=\linewidth]{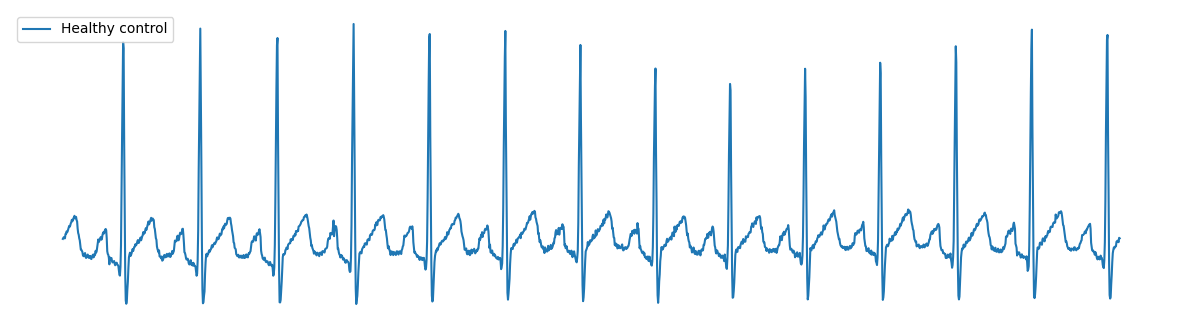}
        \caption{ ECG Plot for Healthy control}
        \label{fig:e}
    \end{subfigure}
    \caption{ECG signal for a representative subjects of different CVDs. The pattern of ECG signal shows clear differences among different CVDs (Figure~\ref{fig:a},~\ref{fig:b},~\ref{fig:c},~\ref{fig:d}) as compared to healthy control (Figure~\ref{fig:e})}
    \label{fig:1}
    
\end{figure}

The Physikalisch-Technische Bundesanstalt (PTB) dataset from the PhysioNet\footnote{Link to dataset: https://www.physionet.org/content/ptbdb/1.0.0/} is employed in this study \cite{[21]}. This dataset comprises 549 recordings of standard 12-lead ECG signals from 290 subjects aged between 17 and 85 years. 
Each record includes 15 channels simultaneously measured signals: the conventional 12 lead (i, ii, iii,  avr, avl,  avf, v1, v2, v3, v4, v5, v6) together with the 3 Frank lead ECGs($v_{x},v_{y},v_{z}$). Each signal is digitized at 1000Hz, with 16-bit resolution over a range of \(\pm 16.384mV\)\cite{[20],[21]}. 
The diagnostic classes in the dataset include Myocardial Infarction, Cardiomyopathy, Bundle Branch Block, Dysrhythmia, Myocardial Hypertrophy, Valvular Heart Disease, Myocarditis, healthy control, and a few miscellaneous subjects. However, due to the limited number of samples, this study excludes data on Myocardial Hypertrophy, Valvular Heart Disease, and Myocarditis.
Thus, in the current study, we utilize ECG recordings of Myocardial Infarction ($MI: n=60$), Bundle Branch Block ($BBB: n=19$), Cardiomyopathy ($CM: n=15$), Dysrhythmia ($DR: n=14$), and healthy control ($HC: n=62$), obtained from the publicly available PhysioNet database.
Several studies have demonstrated that downsampling ECG signals to 500 or 250 Hz yields excellent concordance \cite{new9,new10,new11}. A sampling frequency of 250 Hz is considered acceptable for heart rate variability (HRV) analysis \cite{[14]}. Therefore, in this study, the original ECG signals are downsampled to 250 Hz. 
The downsized ECG signals for various classes of cardiac disorders are shown in Figure~\ref{fig:1}.

\section{PROPOSED METHODOLOGY}
\label{sec3}
\begin{figure}
    \centering
    \includegraphics[width=1\linewidth]{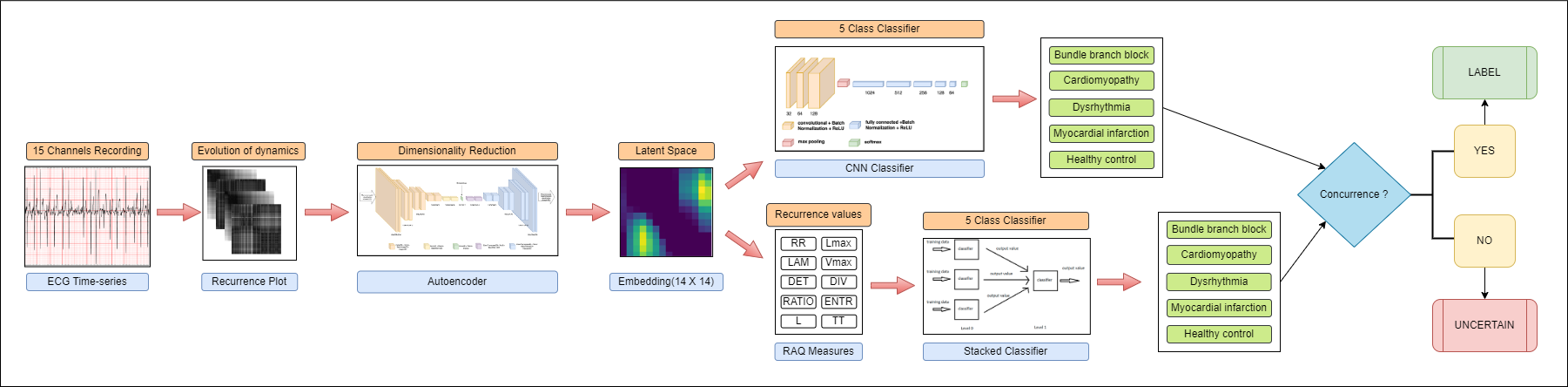}
    \caption{Overview of the proposed methodology}
    \label{fig:2}
\end{figure}

The present study aims to classify various cardiac disorders, including bundle branch block, cardiomyopathy, dysrhythmia, and myocardial infarction, as well as healthy subjects, using latent space embeddings of Recurrence plots and RQA measures. The block diagram in Figure~\ref{fig:2} shows an overview of the proposed methodology.

\subsection{Spatial encoding of ECG using Recurrence plot}

To explore the complexity within each ECG signal, our approach employs nonlinear complexity analysis through Recurrence plots \cite{new1}. This method is highly effective and widely recognized for visualizing and understanding nonlinear time series. The concept of recurrences was first introduced by Henri Poincaré in 1890, with the Poincaré recurrence theorem stating that certain dynamical systems will eventually return to a state that is arbitrarily close to or exactly the same as their initial state after a sufficiently long but finite period. This fundamental property of many dynamic systems is also observed in various natural processes \cite{new2}. In 1987, Eckmann et al. \cite{new3} advanced the visualization of dynamical systems recurrences through Recurrence plots, which were later combined with RQA. RQA is a nonlinear data analysis technique that quantifies the number and duration of recurrences in a dynamical system based on its state space trajectory \cite{new4}. This method has been applied to time series from nonlinear deterministic systems such as financial markets and weather forecasting. For example, Fabretti and Ausloos \cite{new5} used Recurrence plots and RQA to identify critical regimes in financial indices, while Peter Martey Addo et al. \cite{new6} employed these tools to uncover hidden patterns and characterize financial cycles during crises. In neuroscience, Bielski et al. \cite{new7} utilized RQA to parcellate the human amygdala, and Kang et al. \cite{new8} applied RQA to analyze the duration, predictability, and complexity of the default mode network time series in schizophrenia studies. 
In this study, we adapt autoencoder latent space embeddings of Recurrence plots and RQA measures to analyze ECG signals for classifying the considered CVDs, as described in Section~\ref{sec2}. By investigating the distinct patterns revealed in Recurrence plots, we can effectively distinguish between different CVDs, offering valuable insights into the unique dynamics of heart activity for each condition.

\subsubsection{Construction of Recurrence plots}

For a ECG signals \(\{x_1, x_2, \ldots, x_N\}\) consisting of \(N\) timestamps, we create \(K\) state vectors. Each state vector \(\vec{s}_i\) is an \(M\)-dimensional vector defined by:
\begin{equation}
\vec{s}_i = (v_i, v_{i+\tau}, v_{i+2\tau}, \ldots, v_{i+(N-1)\tau})
\label{eq1}
\end{equation}
To apply non-linear analysis, the optimal values of the required parameters, embedding dimension (\(M\)), time lag (\(\tau\)), and number of states (\(K\)), are determined using the traditional Cao's algorithm\cite{[22]}.
The equation~\ref{eq2} is used to construct the data matrix, represented as $\mathcal{D}$ which consists of \( K \) states stacked together. The dimension of $\mathcal{D}$ is \( K \times M \). This matrix $\mathcal{D}$ is utilized to define the Recurrence Matrix ($\mathcal{R}$)\cite{[23]}.

\begin{equation}
\mathcal{D} = 
\begin{bmatrix}
\vec{s}_1 \\
\vec{s}_2 \\
\vdots \\
\vec{s}_n
\end{bmatrix}_{(K \times M)}
\label{eq2}
\end{equation}

In $\mathcal{R}$, the distance between two state vectors is computed using their Euclidean distance, quantifying their dissimilarity. Hence, for a given tuple \((i, j)\), the distance between states \(\vec{s}_i\) and \(\vec{s}_j\) is the \((i,j)\)-th element of the $\mathcal{R}$, given by Equation~\ref{eq3} \cite{[23]}.

\begin{equation}
\mathcal{R}(i, j) = \text{dist}(\vec{s}_i, \vec{s}_j)
\label{eq3}
\end{equation}
i.e., 
\begin{equation}
\mathcal{R} = 
\begin{bmatrix}
\text{dist}(\vec{s}_1, \vec{s}_1) & \cdots & \text{dist}(\vec{s}_1, \vec{s}_K) \\
\text{dist}(\vec{s}_2, \vec{s}_1) & \cdots & \text{dist}(\vec{s}_2, \vec{s}_K) \\
\vdots & \ddots & \vdots \\
\text{dist}(\vec{s}_K, \vec{s}_1) & \cdots & \text{dist}(\vec{s}_K, \vec{s}_K)
\end{bmatrix}_{(K \times K)}
\end{equation}

Eckmann et al.\cite{new3} introduced a method to visualize the recurrence of states \([\vec{s}_i]\) within a phase space. This approach involves projecting high-dimensional phase spaces into two- or three-dimensional spaces for easier visualization. 
The recurrence of a state at time \( x \) at a different time \( y \) is depicted in a two-dimensional square matrix,
with both axes representing time \cite{[24]}. This graphical visualization is known as a Recurrence plot (RP) and is mathematically represented as:
\[ \mathcal{R}_{x,y} = \Theta (\epsilon_x - \| \vec{s}_x - \vec{s}_y \|), \quad \vec{s}_x \in \mathbb{R}^n, \quad x, y = 1, \ldots, N, \]
where 
$\Theta$ is the Heaviside step function, 
$\epsilon_i$ is a threshold distance, and 
$\| \vec{s}_x - \vec{s}_y \|$ is the distance between states 
$\vec{s}_x$ and 
$\vec{s}_y$.

\begin{figure} [H]
    \centering
    \includegraphics[width=1\linewidth]{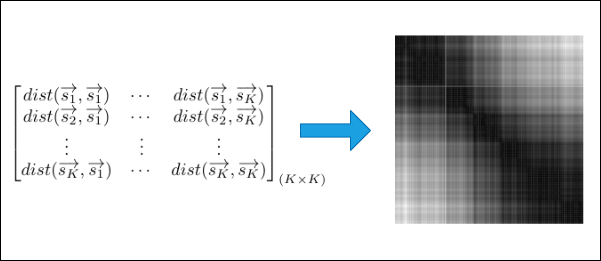}
    \caption{ Recurrence plot visualization by converting an Recurrence matrix of size \(K\times K\) to uniformly-sized (\(224\times224\)) Recurrence image}
    \label{fig:3}
\end{figure}

In this study, the $\mathcal{R}$ matrix of dimensions $K\times K$ is resized to a shape of $224\times 224$ to ensure consistency across all ECG channel signals as well as across all subjects.
The Recurrence plot is shown in Figure~\ref{fig:3}. The information for a specific channel is represented as an gray scale image of size \(1 \times 224 \times 224\). 

\subsection {Autoencoder Architecture}
\label{subsec4}

\begin{figure}[H]
    
    \centering
    \includegraphics[width=1\linewidth]{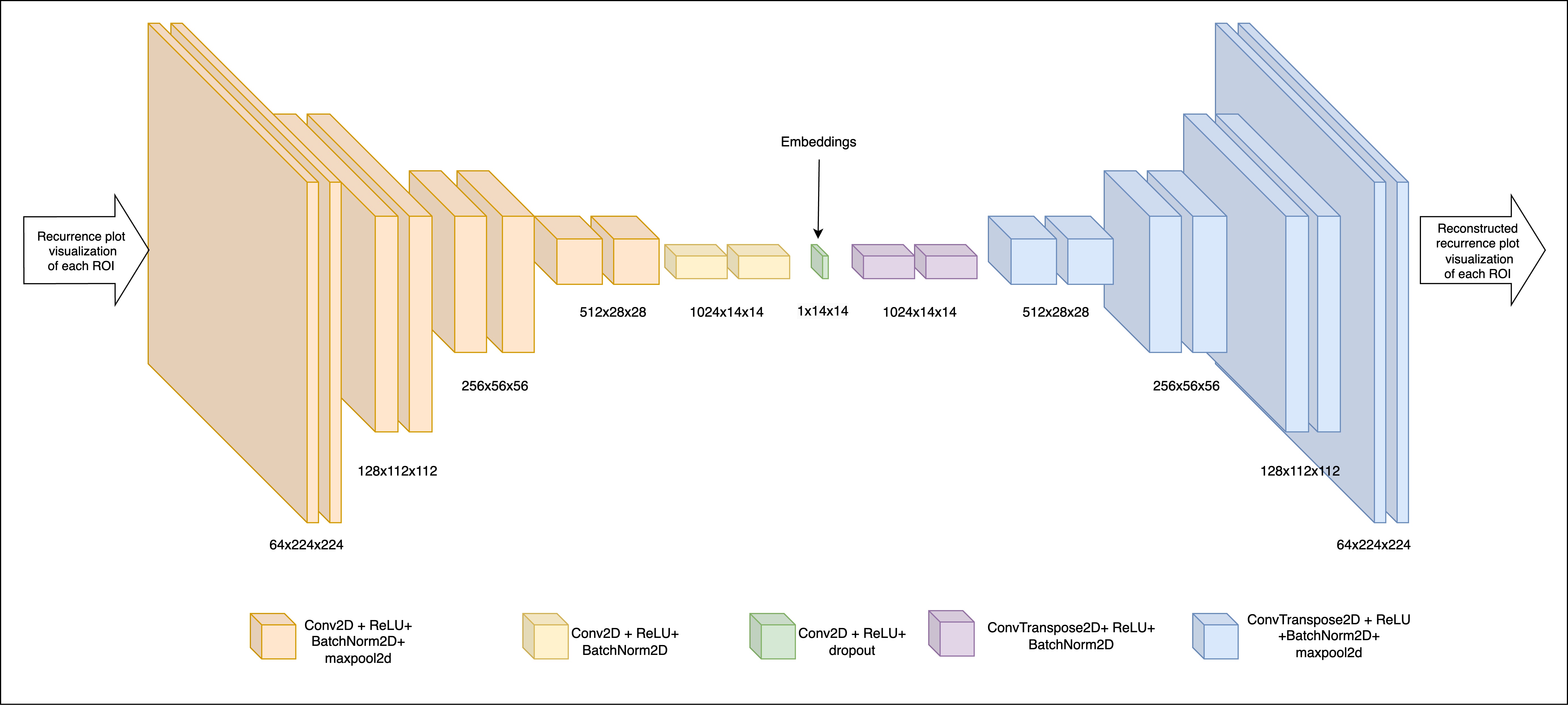}
    \caption{Autoencoder architecture}
    \label{fig:4}
\end{figure}

Each ECG channel is represented as a 2D grayscale image of size $1\times224\times224$. For a subject with $N$ channels, the Recurrence plots are stacked while preserving the channel ordering across all subjects. This results a high-dimensional data point of size $N\times224\times224$. In our study, with 15 ECG channels, each subject is represented as a  $15\times224\times224$ data point. This single high dimensional representation thus efficiently fuses information across all the ECG channels. In order to reduce this dimensionality and capture the essential characteristics, we have used an autoencoder to obtain a feature embedding of size \(14 \times 14\).
As illustrated in Figure~\ref{fig:4}, we utilize a CNN-based autoencoder to obtain the latent space embedding of the Recurrence plots. The autoencoder processes the input Recurrence plots with a size of \(15 \times 224 \times 224\). The convolutional layers of the CNN are designed to capture the essential patterns and features present in the Recurrence plots.
The training of the autoencoder involves optimizing a loss function composed of two terms, as shown in Equation~\ref{eq4}. The first term focuses on reconstruction fidelity, ensuring that the reconstructed output is as close as possible to the original input. The second term introduces a penalty based on the Mean Structural Similarity Index (MSSIM)\cite{[25]}, which helps preserve the structural information within the Recurrence plots.
To train the autoencoder, we used the Adam optimizer, known for its efficiency in handling large datasets and complex models. The training was performed on $2\times T4$ GPUs, which significantly accelerated the process. It took approximately 3.5 hours to train the autoencoder for 1000 epochs. A batch size of 16 was chosen to ensure stable and efficient training, and a train-test split ratio of 0.2 was used to validate the model's performance and avoid overfitting.
The resulting latent space embedding, obtained through this training process, effectively reduces the dimensionality of the original \(15 \times 224 \times 224\) Recurrence plots to a size of \(14 \times 14\), while preserving the critical features necessary for subsequent analysis and classification of the ECG signals.

\begin{equation}
L(x) = |x - \tilde{x}| + (1 - \text{MSSIM}(x, \tilde{x}))
\label{eq4}
\end{equation}

\begin{figure}[H]
    \centering
    \includegraphics[width=1\linewidth]{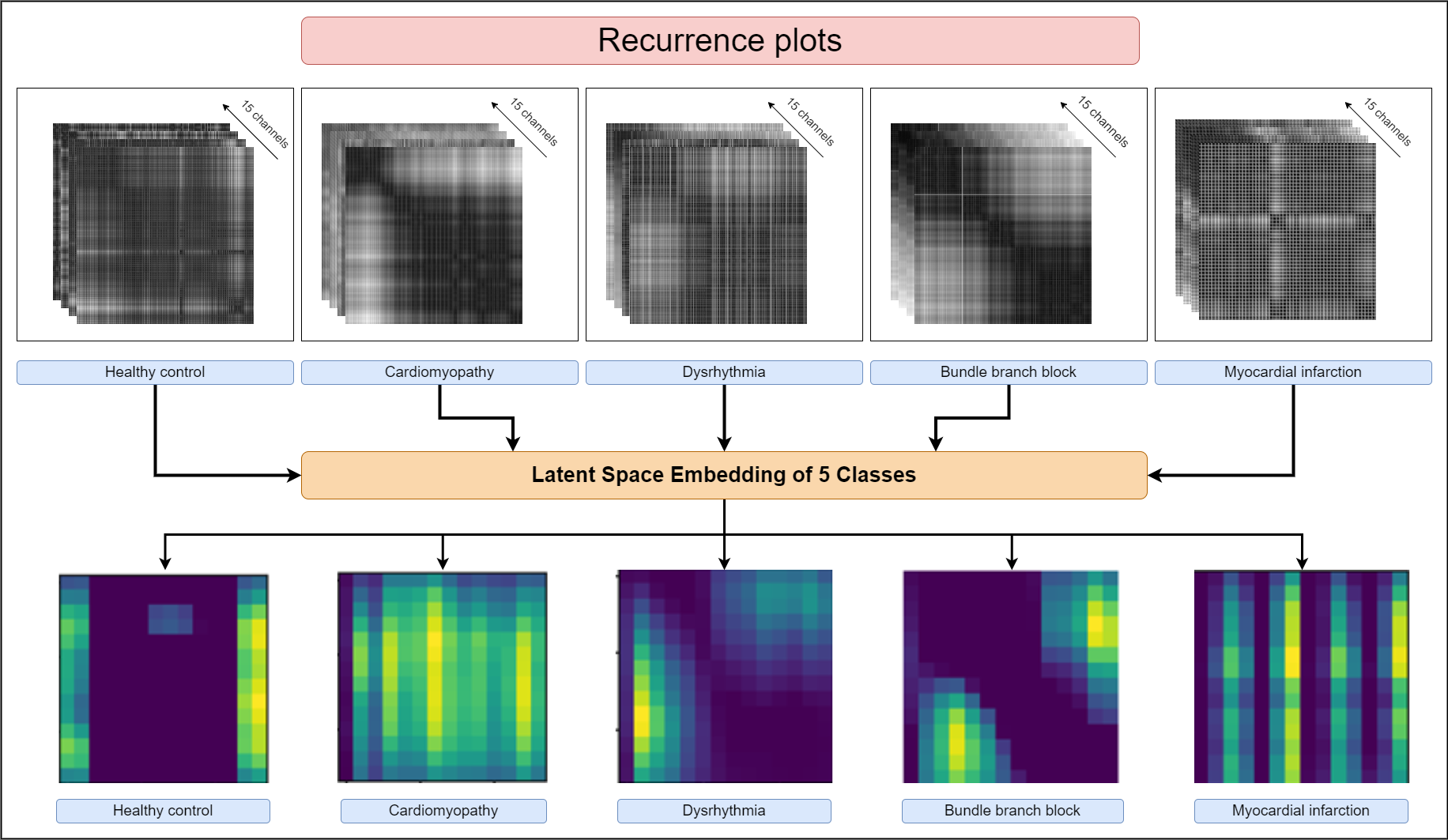}
    \caption{Visualization of autoencoder 2D latent space embedding from Recurrence plots of ECG signal across 15 channels. As seen from this figure, the latent space of the autoencoder showed a clear and distinctive frequency patterns among different CVDs and HC. 
    }
    \label{fig:5}
\end{figure}

\subsection {Latent Space RQA Based Feature Extraction}
\label{subsec5}
RQA is used to quantify the number and duration of recurrences in a dynamical system, based on the trajectory of its state space. In this study, we employ $14\times14$ latent space embeddings, obtained from the autoencoder, to extract RQA features. The RQA values derived from these latent space embeddings are then used for classification purposes. From the latent space embeddings, the current study utilizes a total of 10 RQA features for classification. These features are described in detail below:

\textbf{F1. Recurrence Rate (RR)}: {The percentage of recurrence points in a Recurrence plot (RP) that corresponds to the correlation sum. RR states the recurrence probability of a specific state. Higher value in RR suggests more frequent state recurrences, often associated with more regular or deterministic systems.}
\[RR = \frac{1}{N^2} \sum_{x,y=1}^{N} R_{x,y}\]

\textbf{F2. Determinism (D)}: The percentage of recurrence points forming diagonal lines. \( P(len) \) is the histogram of the lengths \( len \) of the diagonal lines. $D$ provides predictability for the dynamic system.A deterministic process has a Recurrence plot with relatively few single dots and numerous lengthy diagonal lines, whereas white noise has only single dots and very few diagonal lines. High value in $D$ implies the system's behaviour is deterministic, showing predictability and structured patterns over time.
\[D = \frac{\sum_{len=len_{\min}}^{N} len P(len)}{\sum_{len=1}^{N} len P(len)}\]

\textbf{F3. Average Diagonal Line Length $(\bar{d_{L}})$ }: The average length of the diagonal lines. It provides the predictability time of the dynamical system. Longer diagonal lines suggest longer predictable periods, indicating that the system exhibits more extended predictable dynamics.
\[\bar{d_{L}} = \frac{\sum_{len=len_{\min}}^{N} len P(len)}{\sum_{len=len_{\min}}^{N} P(len)}
\]

\textbf{F4. Longest Diagonal Line ($d_{L}{}_{\text{max}}$)}: The length of the longest diagonal line. Longer ($d_{L}{}_{\text{max}}$) values indicate more persistent deterministic behaviour.
\[d_{L}{}_{\text{max}} = \max \{ len_x \;|\; x = 1, \ldots, N_{len} \}\]

\textbf{F5. Entropy (H)}: The Shannon entropy of the probability distribution of the diagonal line lengths \( p(len) \). Entropy represents the complexity of the deterministic structure in the system. Higher entropy indicates more complexity and less predictability, reflecting a mix of different pattern lengths.
\[H = - \sum_{len=len_{\min}}^{N} p(len) \ln p(len)
\]

\textbf{F6. Laminarity (LAM)}: The percentage of recurrence points forming vertical lines.\(p(vl)\) is the histogram of the lengths \(vl\) of the vertical lines. Laminarity gives the value for the amount of laminar phases in the system (intermittency). High value in $LAM$ suggests that the system spends more time in specific states, indicative of intermittent or laminar phases.
\[LAM = \frac{\sum_{vl=vl_{\min}}^{N} vl P(vl)}{\sum_{vl=1}^{N} vl P(vl)}
\]

\textbf{F7. Trapping Time (TT)}: The average length of the vertical lines. It is related to the laminarity time of the dynamical system. Longer trapping times suggest more persistent state trapping.
\[TT = \frac{\sum_{vl=vl_{\min}}^{N} vl P(vl)}{\sum_{vl=vl_{\min}}^{N} P(vl)}
\]

\textbf{F8. Longest Vertical Line ($V_{\text{max}}$)}: The length of the longest vertical line. 
\[V_{\text{max}} = \max \{ vl_x \;|\; x = 1, \ldots, N_{vl} \}\]

\textbf{F9. Divergence (DIV)}: The inverse of $L_{\text{max}}$, related to the KS entropy of the system, i.e., the sum of the positive Lyapunov exponents. The reciprocal of the maximal length of the diagonal lines is an estimator for the positive maximal Lyapunov exponent of the dynamical system as the length of the diagonal lines is related to the time how long segments of the phase space trajectory run parallel or on the divergence behaviour of the trajectories. Longer trapping times suggest more persistent state trapping.
\[DIV = \frac{1}{d_{L}{}_{\text{max}}}
\]

\textbf{F10. RATIO}: It is the ratio between $D$ and $RR$. Higher value in ratio suggest a more deterministic system relative to its recurrence density. 
\[RATIO = N^2 \frac{\sum_{len=len_{\min}}^{N} len P(len)}{\left( \sum_{len=1}^{N} len P(len) \right)^2}
\]

\subsection{Classification of CVDs}
\label{subsec6}

To classify various CVDs, we employ two distinct classifiers: a custom CNN-based classifier and a stacked classifier. Each classifier leverages the latent space feature embeddings derived from the Recurrence plots of ECG signals.

\subsubsection{CNN-Based Classifier}

The first classifier is a custom CNN-based model designed specifically for this task. The input to this CNN classifier is the set of latent space feature embeddings obtained from the Recurrence plots. The architecture of this network includes 3 convolutional layers which are responsible for capturing the essential patterns and features within the latent space embeddings and 6 fully connected layers which are used for prediction, refining the extracted features, and making the final classification.
The network is trained using a cross-entropy loss function, which is well-suited for classification tasks. The Adam optimizer is employed to enhance the training process by efficiently updating the network's weights.

\subsubsection{Stacked Classifier}

The second classifier is a stacked classifier, an ensemble method that combines the strengths of multiple classifiers. The stacked classifier operates as follows: \\
(i) Base Models: The first layer of the stacking ensemble consists of three machine learning-based classifiers: Support Vector Machines (SVM), XGBoost, and RUSBoost.\\
(ii) Meta-Classifier: The second layer, also known as the meta-model, is a logistic regression model. The role of the meta-classifier is to combine the predictions made by the base models to produce the final classification.
Each base model is independently trained on the training dataset. This diversity allows each model to capture different patterns and relationships within the data that might be missed by a single model. After the base models have been trained, their predictions on the training dataset are used as inputs for the meta-classifier. The meta-classifier learns how to best combine these predictions to make the final decision.
The input to the stacked classifier is the RQA-based features of the autoencoder latent space embeddings derived from the Recurrence plots. The RQA values capture the intricate dynamics and structural information of the ECG signals, providing a rich feature set for the classifiers.
By utilizing both a custom CNN-based classifier and a sophisticated stacked classifier, we leverage the strengths of deep learning and ensemble methods to achieve robust and accurate classification of cardiac disorders. The custom CNN excels in capturing spatial features from the Recurrence plots, while the stacked classifier benefits from the diverse perspectives of multiple machine learning models, combined into a powerful meta-classifier. This dual approach ensures a comprehensive analysis and enhances the reliability of our classification system.

\begin{figure}[H]
    \centering
    \includegraphics[width=0.9\linewidth]{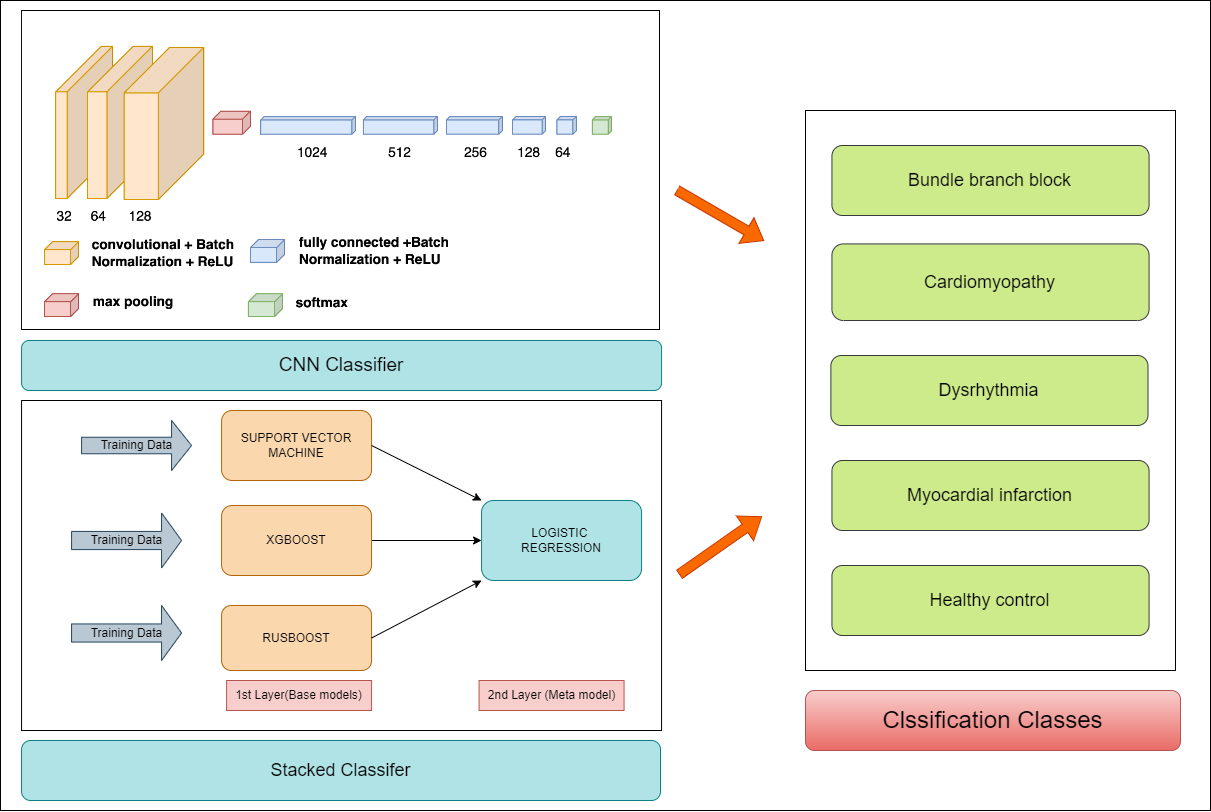}
    \caption{Architecture of CNN and Stacked classifier used in this study}
    \label{fig:6}
\end{figure}

\section{EXPERIMENTAL RESULTS}
\label{sec4}

\begin{figure}
    \centering
    \includegraphics[width=1\linewidth]{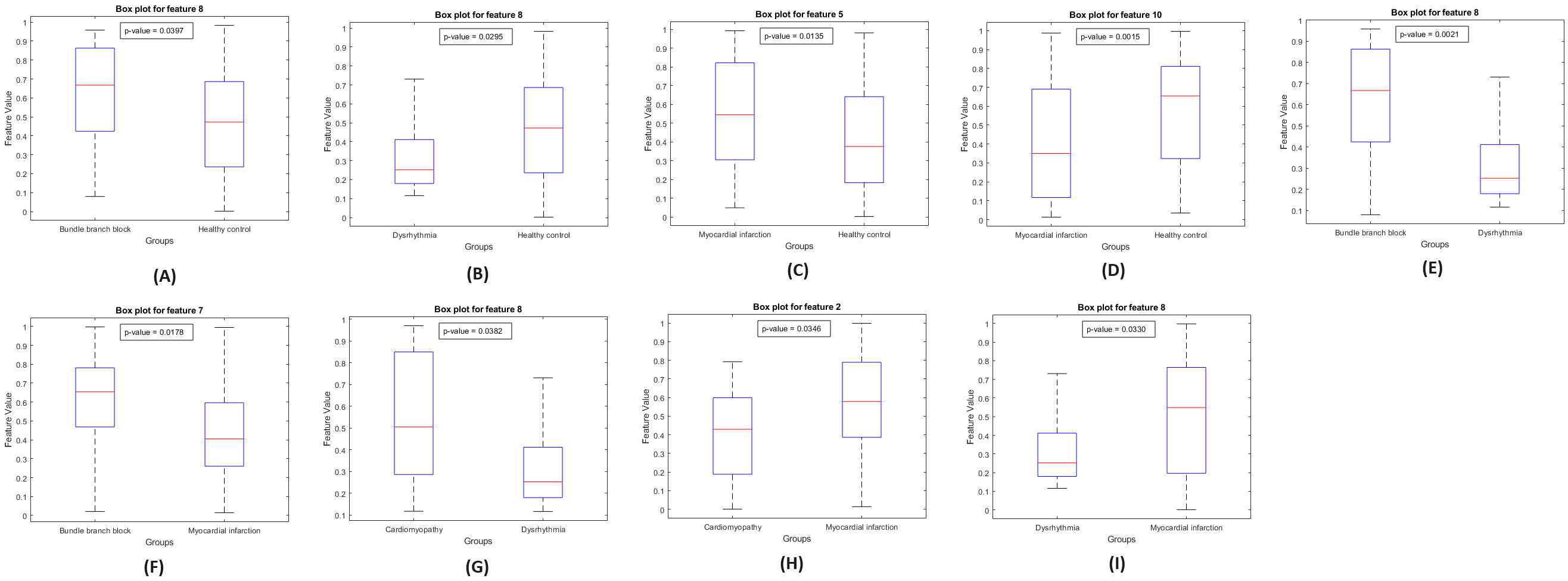}\textbf{}
    \caption{The box plot of 9 RQA features that showed statistically significant differences between study groups (summarized in Table~\ref{tab:stat_test_results}) }
    \label{fig:stats}
\end{figure}

\begin{figure}[H]
    \centering
    \begin{subfigure}[b]{0.45\linewidth}
        \centering
        \includegraphics[width=\linewidth]{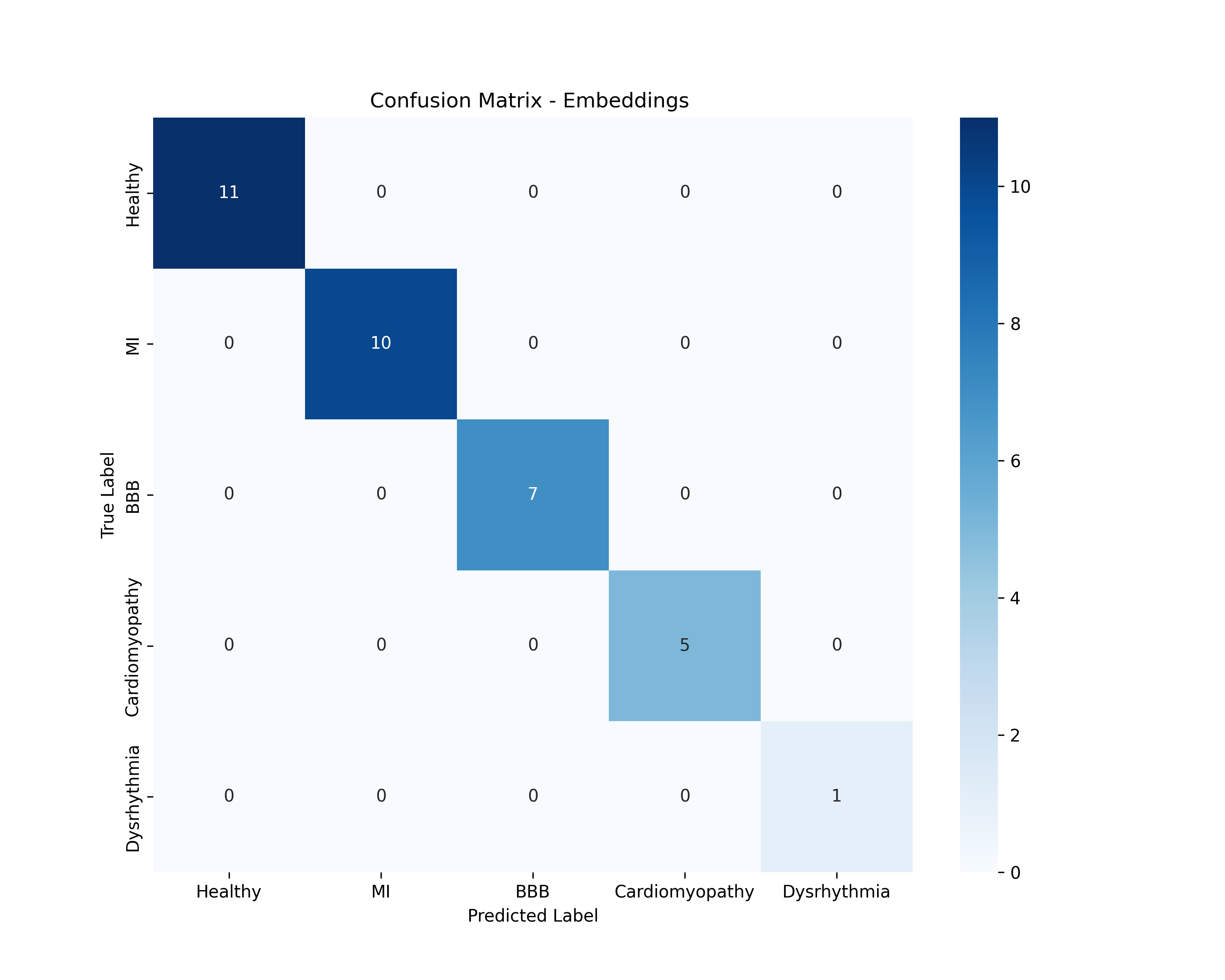}
        \caption{Classification performance using latent space embedding of Recurrence plot}
        \label{fig:8.a}
    \end{subfigure}
    \hfill
    \begin{subfigure}[b]{0.45\linewidth}
        \centering
        \includegraphics[width=\linewidth]{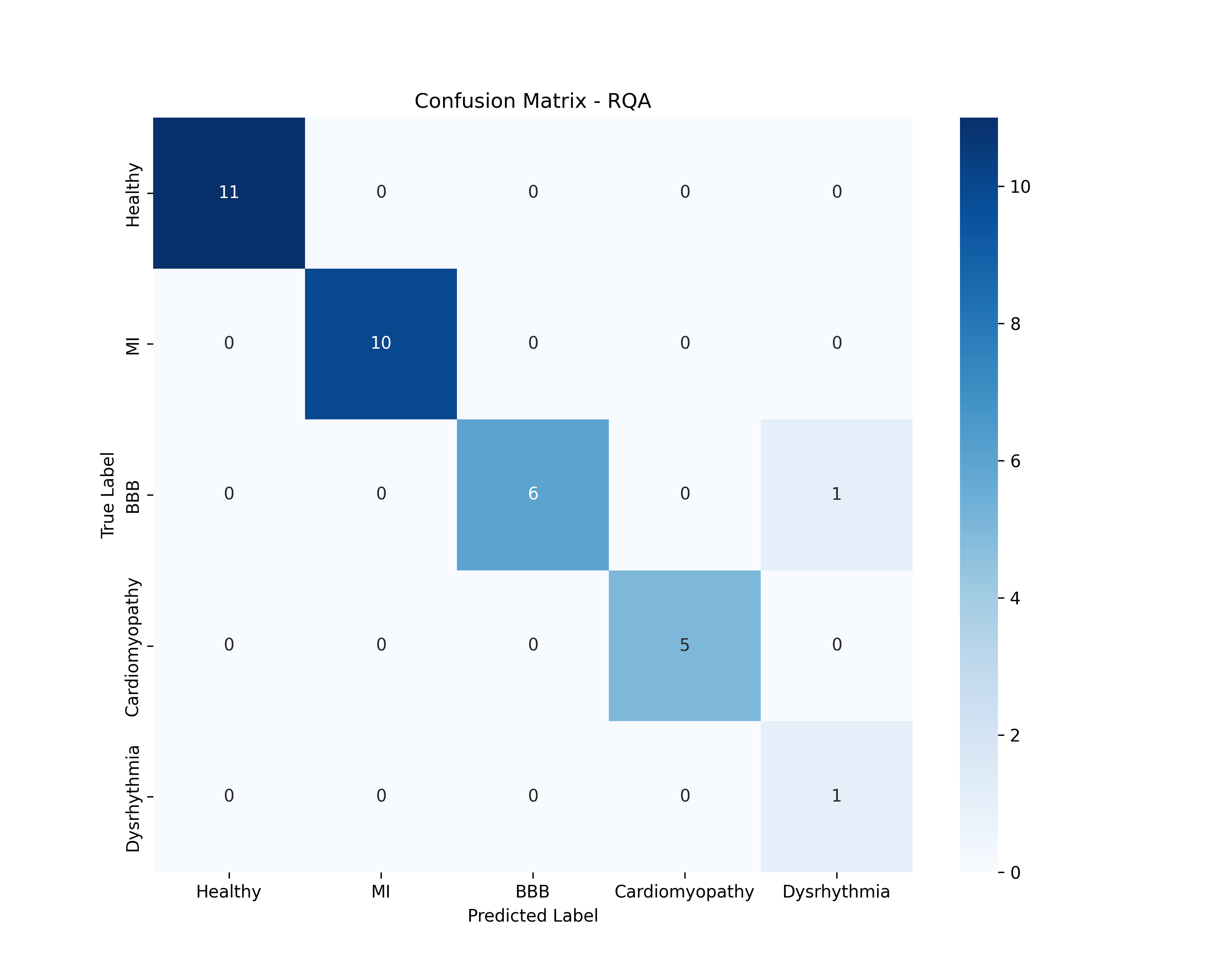}
        \caption{Classification performance using RQA features of Recurrence plot latent space embedding }
        \label{fig:8.b}
    \end{subfigure}
    \caption{Heat map of confusion matrices obtained from the study.}
    \label{fig:8}
\end{figure}

\begin{figure}[H]
    \centering
    \begin{subfigure}[b]{0.45\linewidth}
        \centering
        \includegraphics[width=\linewidth]{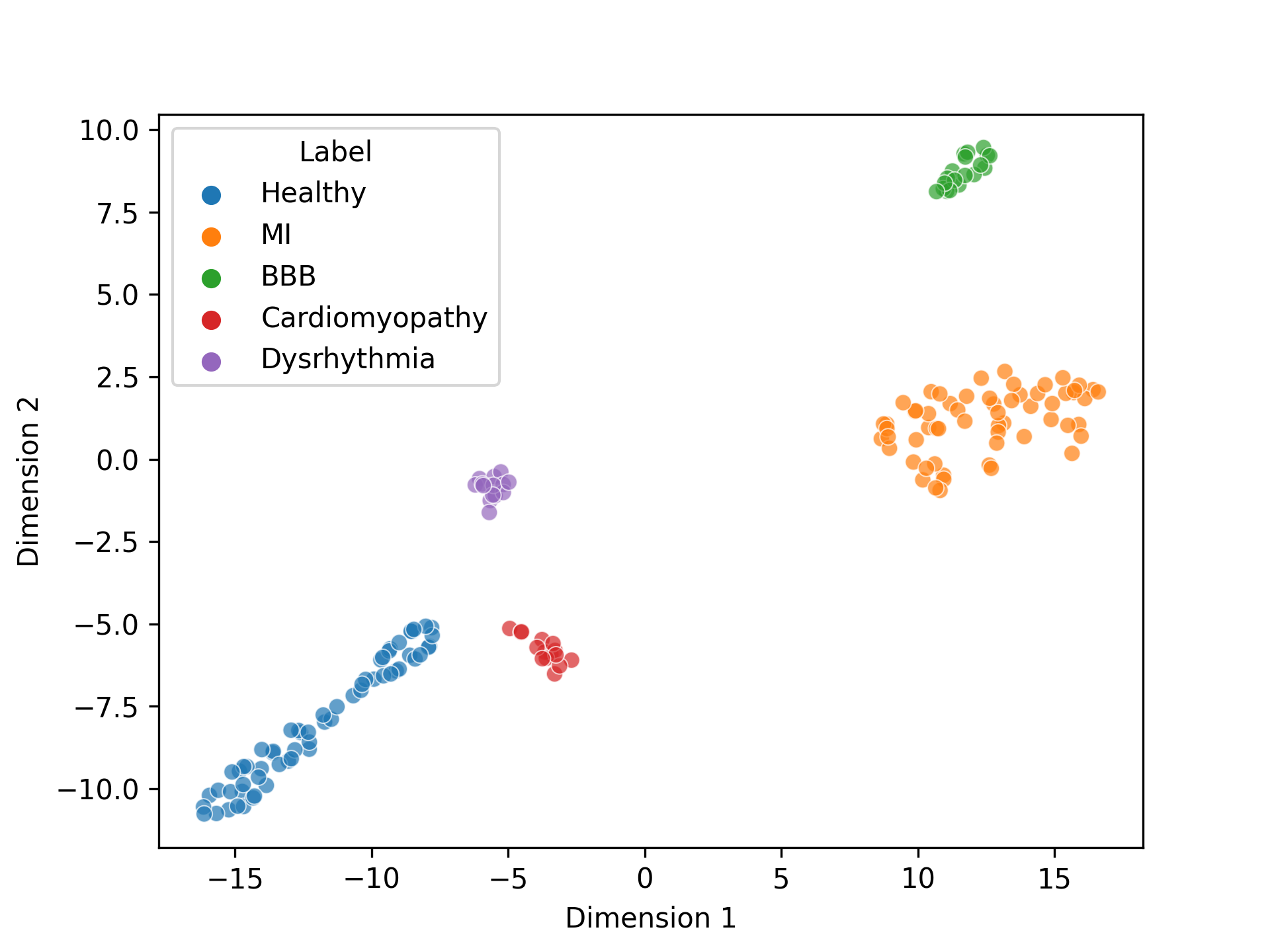}
        \caption{t-SNE visualization of latent space embeddings of Recurrence plot}
        \label{fig:9.a}
    \end{subfigure}
    \hfill
    \begin{subfigure}[b]{0.45\linewidth}
        \centering
        \includegraphics[width=\linewidth]{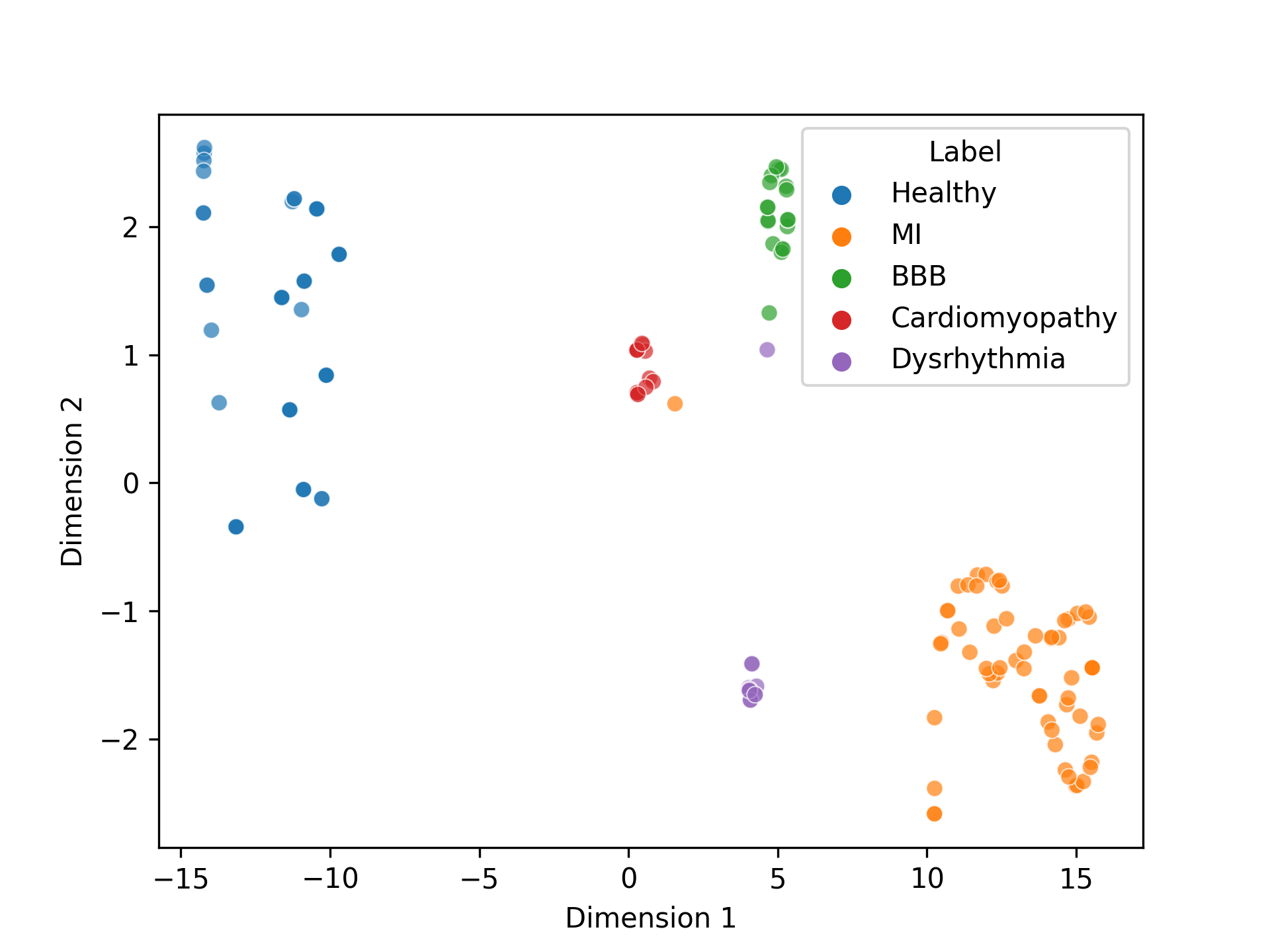}
        \caption{ t-SNE visualization of RQA features of latent space embedding}
        \label{fig:9.b}
    \end{subfigure}
    \caption{The t-SNE visualization on test data illustrate inter-class separability among different CVDs and HC achieved using the proposed method.}
    \label{fig:9}
\end{figure}

The present study combines deep learning, ensemble methods, and rigorous statistical analysis to ensure robust and accurate classification of various CVDs. 
The proposed methodology processes ECG signals by first converting each channel into 2D Recurrence plots of size \(224 \times 224\), resulting in a high-dimensional input of size \(15 \times 224 \times 224\) for each recording. These images are fed into an autoencoder to reduce dimensionality and extract essential features, producing feature embeddings of size \(14 \times 14\). These embeddings are used to extract RQA features. Wilcoxon Rank-Sum test at 95\% confidence interval is conducted to identify RQA features where the differences are statistically significant ($p<0.05$). The result of this statistical test is presented in Table~\ref{tab:stat_test_results}. Figure~\ref{fig:stats} shows the box-plot of the obtained 9 RQA features that showed statistically significant differences between study groups. 
For classification purposes, the latent space of Recurrence plots and RQA features were used as inputs to two different classifiers: (1) a custom CNN-based classifier, which employed the latent space of Recurrence plots, achieved a peak accuracy of 100\%, and (2) a stacked classifier that utilized RQA features reached a peak accuracy of 97.05\%. The five-class classification results for four cardiac disorders and healthy controls are detailed in Table~\ref{table:1}, with heat map of confusion matrices in Figures~\ref{fig:8.a} and \ref{fig:8.b}. The t-SNE plots in Figure~\ref{fig:9} show well-separated clusters for both classifiers, highlighting the efficacy of the proposed technique.

\begin{table}[H]
\centering
\begin{tabular}{ |c|c|c|c|c|c|c|c|c|c|c|c| } 
\hline
\textbf{Comp.} & \textbf{F1} & \textbf{F2} & \textbf{F3} & \textbf{F4} & \textbf{F5} & \textbf{F6} & \textbf{F7} & \textbf{F8} & \textbf{F9} & \textbf{F10}  \\
\hline
HC-BBB & ns& ns & ns &ns & ns & ns & ns & 0.039 &ns & ns  \\
HC-CM & ns& ns & ns & ns & ns &ns &ns &ns &ns & ns  \\
HC-DR & ns & ns & ns & ns & ns & ns &ns & 0.029 & ns & ns \\
HC-MI & ns & ns & ns & ns & 0.013 & ns & ns & ns & ns & 0.001  \\
\hline
BBB-CM & ns & ns & ns & ns & ns &ns &ns &ns &ns & ns  \\
BBB-DR & ns & ns & ns & ns & ns &ns &ns & 0.002&ns & ns  \\
BBB-MI & ns & ns & ns & ns & ns &ns & 0.017 & ns & ns & ns  \\
\hline
CM-DR & ns & ns & ns & ns & ns &ns &ns & 0.038 &ns &ns   \\
CM-MI & ns & 0.034 & ns & ns & ns & ns &ns &ns &ns &ns  \\
\hline
DR-MI & ns & ns & ns & ns & ns &ns &ns & 0.033 &ns &ns  \\
\hline
\end{tabular}
\caption{Results of Wilcoxon rank-sum test, conducted at 95\% C.I., for pairwise comparison between study groups (`ns' indicates the difference is not statistically significant). }
\label{tab:stat_test_results}
\end{table}

\begin{table}[H]
\centering
\begin{tabular}{ |c|c|c|c|c|c| } 
\hline
\textbf{Models} & \textbf{Classes} & \textbf{Precision} & \textbf{Recall} & \textbf{F1-score} & \textbf{Accuracy}\\
\hline
\multirow{3}{4em}{Stacked classifier} & HC & 1.00 & 1.00 & 1.00&\\ 
& MI & 1.00 & 1.00 & 1.00 &\\ 
& BBB & 1.00 & 0.86 & 0.92 & 97.05\% \\
& CM & 1.00 & 1.00 & 1.00 & \\ 
& DR & 0.50 & 1.00 & 0.67 & \\ 
\hline
\multirow{3}{4em}{CNN classifier} & HC & 1.00 & 1.00 & 1.00 &\\ 
& MI & 1.00 & 1.00 & 1.00 &\\ 
& BBB & 1.00 & 1.00 & 1.00 & 100\% \\ 
& CM & 1.00 & 1.00 & 1.00 & \\ 
& DR & 1.00 & 1.00 & 1.00 & \\ 
\hline
\end{tabular}
\caption{The five-class classification performance of the proposed methodology is evaluated using two models: (1) the stacked classifier utilizes RQA features, while (2) the CNN utilizes latent space embeddings of Recurrence plot.}
\label{table:1}
\end{table}

\section{DISCUSSION}
\label{sec5}

Recurrence plots are traditionally used to infer structural characteristics, detect chaos, and evaluate predictability or unpredictability within time series data. In the context of studying various CVDs, recurrence plots reveal the complex patterns of recurrence within ECG signals, capturing subtle and intricate dynamical behaviors specific to each condition. They are particularly useful as they visualize and quantify the non-linear and non-stationary aspects of ECG signals, aspects often overlooked by traditional linear analysis methods. By uncovering hidden structures, chaos, and predictability within the ECG data, recurrence plots offer a deeper understanding of cardiac dynamics, enhancing the ability to differentiate between different CVDs. While traditional methods focus on analyzing recurrence plots through individual RQA, our approach introduces novelty by combining feature embeddings of these plots with RQA measures derived from the embeddings. This study seeks to develop a more effective technique for classifying different CVDs using ECG data by integrating Recurrence plots and RQA measures derived from autoencoder latent space embeddings of ECG signals.
The hypothesis of the study is that the dynamics of ECG signals differ in the presence of cardiac abnormalities. We showed that transforming 1D ECG signals into 2D Recurrence plots, followed by extracting RQA measures from autoencoder latent space embeddings, provides a more informative and discriminative feature set for accurate classification as compared to traditional methods. Furthermore, the proposed study demonstrate superior performance across different types of classifiers, including CNN-based classifiers and stacked ensemble classifiers, thereby validating the robustness and generalizability of the approach.
The latent space of autoencoder (Figure~\ref{fig:5}) reveals distinct frequency patterns that could be instrumental in interpreting the dynamics of CVDs. Analyzing these patterns in frequency space offers valuable insights into how these conditions evolve and change over time. Future studies could leverage this frequency information of latent space for better interpretability, to enhance diagnostic accuracy, monitor disease progression, and to develop predictive models. 
The t-SNE plots visually demonstrate the effectiveness of our method. Figure~\ref{fig:9.a} showcases the separability of the feature space spanned by the latent space embeddings, while Figure~\ref{fig:9.b} displays the separability of the feature space defined by the RQA measures of these embeddings. This visual representation underscores the efficacy of our approach in distinguishing between different cardiac conditions.
The recent state-of-the-art classification methods, as detailed in Table~\ref{table:2}, have utilized different sophisticated techniques on the same dataset. These methods include the use of CNNs with focal loss, entropy features, and few-shot learning frameworks. 
In contrast, our method significantly outperforms these existing models. By employing latent space embeddings derived from Recurrence plots, we achieve a perfect classification accuracy of 100\%. Additionally, when using the RQA measures of these embeddings, our method attains high accuracy of 97.05\%. 
This substantial improvement in performance is noteworthy. Specifically, our method surpasses the current models by a margin of 4\% when using autoencoder latent space embeddings and by 1\% when utilizing RQA measures of these embeddings. These results highlight the robustness and effectiveness of our approach in classifying various CVDs.
The proposed methodology leverages the strengths of Recurrence plots and RQA in a novel manner, demonstrating superior performance compared to state-of-the-art methods. This advancement is critical for improving the accuracy and reliability of ECG classification, thereby contributing to better diagnostic tools in cardiology. Future work could explore integrating additional non-linear analysis techniques and advanced deep learning architectures, such as transformer models or graph neural networks, to enhance feature extraction and classification accuracy. Expanding the dataset to include a wider range of cardiac conditions and diverse patient demographics, along with implementing real-time processing capabilities, would help validate the generalizability of the proposed approach and practical applicability in clinical settings.

\begin{table}[H]
\centering
\begin{tabular}{ |@{}c@{}|@{}c@{}|@{}c@{}|@{}c@{}|@{}c@{}|@{}c@{}| } 
\hline
\textbf{REF} & \textbf{Year} & \textbf{Subject} & \textbf{Methods} & \textbf{Classifier} & \textbf{Accuracy} \\
\hline
\multirow{1}{*}{\cite{[8]}} & \multirow{1}{*}{2021} & HC, MI, & Data & CNN & \multirow{1}{*}{76.5\%} \\ 
&&CD, STTC, & filtering& with entropy & \\
&& and HYP  & & features &\\
\hline
\multirow{1}{*}{\cite{[9]}} & \multirow{1}{*}{2022} & HC, MI, & QRS  & Few   & \multirow{1}{*}{93.2\%} \\ 
&&CD, STTC, & Extracting& Short &\\
&& and HYP  & & Learning&\\
\hline
\multirow{1}{*}{\cite{[26]}} & \multirow{1}{*}{2021} & AR and & Raw & CNN & \multirow{1}{*}{93.14\%} \\ 
&& HC & Input & NCBAM & \\
\hline
\multirow{1}{*}{\cite{[27]}} & \multirow{1}{*}{2017} & MI and  & Beat  & ML-CNN & \multirow{1}{*}{96\%} \\ 
&&HC & segmentation & &\\
\hline
\multirow{1}{*}{\cite{[28]}} & \multirow{1}{*}{2022} & MI and & Normalization,  & Random  & \multirow{1}{*}{74\%} \\ 
&&HC & segmentation  &  forest &\\
\hline
\hline
\textbf{RP Based} & \multirow{1}{*}{\textbf{2024}} & \textbf{HC,}  & \textbf{Recurrence} & \textbf{CNN} & {\textbf{100\%}} \\
\textbf{Method}& &\textbf{BBB, MI}& \textbf{Plot} & \textbf{Classifier} & \\
\cline{4-6}
&&\textbf{CM, and}&\textbf{RQA } & \textbf{Stacked } & {\textbf{97.05\%}} \\
&&\textbf{DR}& \textbf{Mesures} &  \textbf{Classifier}&  \\
\hline
\end{tabular}
\caption{Comparison of the proposed study with state-of-the-art methods that used the same PTB dataset from PhysioNet.}
\label{table:2}
\end{table}

\section{CONCLUSION}
\label{sec6}
The current study examines the applicability of non-linear analysis of ECG signals using Recurrence plots, autoencoder latent space embedding, and RQA measures in order to detect various cardiac disorders. We hypothesize that ECG signal dynamics vary with the presence of cardiac abnormalities. Low-dimensional feature embeddings offer a fusion of information across multi-channel ECG data, enabling efficient processing for classification. The differences in latent space embedding patterns of Recurrence plot and RQA features among various CVDs highlight the effectiveness of the proposed approach, as evidenced by the promising accuracy.




\begin{thebibliography}{00}


\bibitem{[1]}
Mensah G.A., Roth G.A., Fuster V. "The global burden of cardiovascular diseases and risk factors: 2020 and beyond". J Am Coll Cardiol 2019;74:2529-2532.
\bibitem{[2]}
Roth G.A., Mensah G.A., Fuster V. "The global burden of cardiovascular diseases and risks: a compass for global action". J Am Coll Cardiol 2020;76:2980-2981.
\bibitem{[3]}
Rotman M, Triebwasser JH. A clinical and follow-up study of right and left bundle branch block. Circulation. 1975 Mar;51(3):477-84.
\bibitem{[4]}
Yang H. Multiscale recurrence quantification analysis of spatial cardiac vectorcardiogram signals. IEEE Trans Biomed Eng. 2011 Feb;58(2):339-47. doi: 10.1109/TBME.2010.2063704. Epub 2010 Aug 5. PMID: 20693104.
\bibitem{[5]}
U.R. Acharya, J.S. Suri, J.A.E. Spaan, S.M. Krishnan, Advances in Cardiac Signal Processing, Springer-Verlag Berlin Heidelberg, New York, 2007.
\bibitem{[6]}
A.L. Goldberger, Clinical Electrocardiography: a Simplified Approach, Mosby, St.Louis, MO, USA, 2012.
\bibitem{[7]}
Hammad, M., Alkinani, M.H., Gupta, B.B. et al. Myocardial infarction detection based on deep neural network on imbalanced data. Multimedia Systems 28, 1373–1385 (2022). https://doi.org/10.1007/s00530-020-00728-8.
\bibitem{[8]}
Śmigiel S, Pałczyński K, Ledziński D. ECG Signal Classification Using Deep Learning Techniques Based on the PTB-XL Dataset. Entropy (Basel). 2021 Aug 28;23(9):1121. doi: 10.3390/e23091121. PMID: 34573746; PMCID: PMC8469424.
\bibitem{[9]}
 Pałczyński K, Śmigiel S, Ledziński D, Bujnowski S. Study of the Few-Shot Learning for ECG Classification Based on the PTB-XL Dataset. Sensors (Basel). 2022 Jan 25;22(3):904. doi: 10.3390/s22030904. PMID: 35161650; PMCID: PMC8839938.

\bibitem{[10]}
T.J. Jun, H.M. Nguyen, D. Kang, D. Kim, D. Kim, Y.H. Kim, ECG Arrhythmia
Classification Using a 2-D Convolutional Neural Network, arXiv preprint arXiv:
1804.06812, 2018.
\bibitem{[11]}
S. Anwar, K. Hwang, W. Sung, Fixed point optimization of deep convolutional
neural networks for object recognition, 2015 IEEE International Conference on
Acoustics, Speech, and Signal Processing (ICASSP) (2015) 1131–1135. April.
\bibitem{[12]}
M. Liang, X. Hu, Recurrent convolutional neural network for object recognition, In
Proceedings of the IEEE Conference on Computer Vision and Pattern Recognition
(2015) 3367–3375.
\bibitem{[13]}
T. Ishii, R. Nakamura, H. Nakada, Y. Mochizuki, H. Ishikawa, Surface object
recognition with CNN and SVM in Landsat 8 images, 14th IAPR International
Conference on Machine Vision Applications (MVA) (2015) 341–344. May. 2015,
IEEE.
\bibitem{[14]}
Kwon O, Jeong J, Kim HB, Kwon IH, Park SY, Kim JE, Choi Y. Electrocardiogram Sampling Frequency Range Acceptable for Heart Rate Variability Analysis. Healthc Inform Res. 2018 Jul;24(3):198-206. doi: 10.4258/hir.2018.24.3.198. Epub 2018 Jul 31. PMID: 30109153; PMCID: PMC6085204.
\bibitem{[15]}
Zhiguang Wang and Tim Oates, “Imaging time-series to improve classification and imputation,” arXiv preprint arXiv:1506.00327, 2015.
\bibitem{[16]}
Norbert Marwan, MCarmenRomano,MarcoThiel, andJ¨urgen Kurths,
 “Recurrence plots for the analysis of complex systems,” Physics reports, vol. 438, no. 5-6, pp. 237–329, 2007.
\bibitem{[17]}
Zhang H, Liu C, Zhang Z, Xing Y, Liu X, Dong R, He Y, Xia L, Liu F. Recurrence plot-Based Approach for Cardiac Arrhythmia Classification Using Inception-ResNet-v2. Front Physiol. 2021 May 17;12:648950. doi: 10.3389/fphys.2021.648950. PMID: 34079470; PMCID: PMC8165394.
\bibitem{[18]}
Zbilut JP, Thomasson N, Webber CL. Recurrence quantification analysis as a tool for nonlinear exploration of nonstationary cardiac signals. Med Eng Phys. 2002 Jan;24(1):53-60. doi: 10.1016/s1350-4533(01)00112-6. PMID: 11891140.
\bibitem{[19]}
Yafei Kang, Youming Zhang, Kexin Huang, and Zhenhong Wang, “Re
currence quantification analysis of periodic dynamics in the default
 mode network in first-episode drug-na¨ ıve schizophrenia,” Psychiatry
 Research: Neuroimaging, vol. 329, pp. 111583, 2023.
\bibitem{[20]}
Bousseljot R, Kreiseler D, Schnabel, A. Nutzung der EKG-Signaldatenbank CARDIODAT der PTB über das Internet. Biomedizinische Technik, Band 40, Ergänzungsband 1 (1995) S 317.

\bibitem{[21]}
Goldberger, A., Amaral, L., Glass, L., Hausdorff, J., Ivanov, P. C., Mark, R., ... \& Stanley, H. E. (2000). PhysioBank, PhysioToolkit, and PhysioNet: Components of a new research resource for complex physiologic signals. Circulation [Online]. 101 (23), pp. e215–e220.
\href{https://physionet.org/content/ptbdb/1.0.0/}{Dataset}

\bibitem{[22]}
Liangyue Cao, “Determining minimum embedding dimension from
 scalar time series,” in Modelling and Forecasting Financial Data: Techniques of Nonlinear Dynamics, pp. 43–60. Springer, 2002.

\bibitem{[23]}
Ninad Aithal and Chakka Sai Pradeep and Neelam Sinha. MCI Detection using fMRI time series embeddings of Recurrence plots. arXiv. 2023;2311.18265.doi: 10.48550/arXiv.2311.18265
\bibitem{[24]}
Eckmann, J.-P., Kamphors, S. O., and Ruell, D. (1987). Recurrence plots of dynamical systems. Europhys. Lett. 4, 973–977.

\bibitem{[25]}
 Zhou Wang, Alan C Bovik, Hamid R Sheikh, and Eero P Simoncelli, “Image quality assessment: from error visibility to structural similarity,” IEEE transactions on image processing, vol. 13, no. 4, pp. 600-612, 2004.

\bibitem{[26]}
 Wang J, Qiao X, Liu C, Wang X, Liu Y, Yao L, Zhang H. Automated ECG classification using a non-local convolutional block attention module. Comput Methods Programs Biomed. 2021 May;203:106006. doi: 10.1016/j.cmpb.2021.106006. Epub 2021 Feb 27. PMID: 33735660.
\bibitem{[27]}
Liu W, Zhang M, Zhang Y, Liao Y, Huang Q, Chang S, Wang H, He J. Real-Time Multilead Convolutional Neural Network for Myocardial Infarction Detection. IEEE J Biomed Health Inform. 2018 Sep;22(5):1434-1444. doi: 10.1109/JBHI.2017.2771768. Epub 2017 Nov 10. PMID: 29990164.
\bibitem{[28]}
Sraitih M, Jabrane Y, Hajjam El Hassani A. A Robustness Evaluation of Machine Learning Algorithms for ECG Myocardial Infarction Detection. J Clin Med. 2022 Aug 23;11(17):4935. doi: 10.3390/jcm11174935. PMID: 36078865; PMCID: PMC9456488.
\bibitem{deb1}
V. K. Kancharala, D. Bhattacharya and N. Sinha, "Spatial Encoding of BOLD fMRI Time Series for Categorizing Static Images Across Visual Datasets: A Pilot Study on Human Vision," TENCON 2023 - 2023 IEEE Region 10 Conference (TENCON), Chiang Mai, Thailand, 2023, pp. 1117-1122, doi: 10.1109/TENCON58879.2023.10322476.
\bibitem{deb2}
Ammu R., Debanjali Bhattacharya, Ameiy Acharya, Ninad Aithal, Neelam Sinha. Multi-scale fMRI time series analysis for understanding neurodegeneration in MCI. arXiv:2402.02811 [cs.CV]. https://doi.org/10.48550/arXiv.2402.02811

\bibitem{new1}
Cao, L. Determining minimum embedding dimension from scalar time series. In Modelling and Forecasting Financial Data: Techniques of Nonlinear Dynamics, 43–60 (Springer, 2002).
\bibitem{new2}
Marwan N, S. P., Kurths J. Generalized Recurrence plot analysis for spatial data. Phys Lett A. 360, 545–551 (2007).
\bibitem{new3}
Eckmann, J.-P., Kamphorst, S. O., Ruelle, D. et al. Recurrence plots of dynamical systems. World Sci. Ser. on Nonlinear Sci. Ser. A 16, 441–446 (1995).
\bibitem{new4}
Webber Jr, C. L. and Zbilut, J. P. Recurrence quantification analysis of nonlinear dynamical systems. Tutorials contemporary nonlinear methods for behavioral sciences 94, 26–94 (2005).
\bibitem{new5}
Fabretti, A. and Ausloos, M. Recurrence plot and recurrence quantification analysis techniques for detecting a critical regime examples from financial market inidices. Int. J. Mod. Phys. C 16, 671–706 (2005).
\bibitem{new6}
Addo, P. M., Billio, M. and Guégan, D. Nonlinear dynamics and Recurrence plots for detecting financial crisis. The North Am. J. Econ. Finance 26, 416–435, DOI: https://doi.org/10.1016/j.najef.2013.02.014 (2013).
\bibitem{new7}
Bielski, K., Adamus, S. et.al. Parcellation of the human amygdala using recurrence quantification analysis. Neuroimage 227, 117644 (2021).
\bibitem{new8}
Kang, Y., Zhang, Y., Huang, K. and Wang, Z. Recurrence quantification analysis of periodic dynamics in the default mode network in first-episode drug-naïve schizophrenia. Psychiatry Res. Neuroimaging 329, 111583 (2023).
\bibitem{deb3}
S. K. Kopparapu, D. Bhattacharya and N. Sinha, "Spatial Encoding of EEG Brain Wave Signals to Predict Student’s Mental State During E-Learning," 2023 IEEE 33rd International Workshop on Machine Learning for Signal Processing (MLSP), Rome, Italy, 2023, pp. 1-6, doi: 10.1109/MLSP55844.2023.10285955.
\bibitem{new9}
Ellis RJ, Zhu B, Koenig J, Thayer JF, Wang Y. A careful look at ECG sampling frequency and R-peak interpolation on short-term measures of heart rate variability. Physiol Meas. 2015 Sep;36(9):1827-52. doi: 10.1088/0967-3334/36/9/1827. Epub 2015 Aug 3. PMID: 26234196.

\bibitem{new10}
Khayatzadeh M, Zhang X, Tan J, Liew WS, Lian Y. A 0.7-V 17.4- $\mu$W 3-lead wireless ECG SoC. IEEE Trans Biomed Circuits Syst. 2013 Oct;7(5):583-92. doi: 10.1109/TBCAS.2013.2279398. Epub 2013 Sep 23. PMID: 24108477.

\bibitem {new11}
Argüello Prada EJ, Paredes Higinio A. A low-complexity PPG pulse detection method for accurate estimation of the pulse rate variability (PRV) during sudden decreases in the signal amplitude. Physiol Meas. 2020 Apr 16;41(3):035001. doi: 10.1088/1361-6579/ab7878. PMID: 32079008.



\end{thebibliography}



\end{document}